\documentclass[11pt]{article}

\usepackage[margin=1in]{geometry}
\usepackage{amsmath,amssymb,amsthm}
\usepackage[hidelinks]{hyperref}

\newcommand{\authorname}{Alexis Goujon}
\newcommand{\repourl}{https://github.com/axgoujon/LipschitzUniversality}

\newtheorem{theorem}{Theorem}
\newtheorem{corollary}[theorem]{Corollary}
\newtheorem{lemma}[theorem]{Lemma}
\theoremstyle{definition}
\newtheorem{definition}[theorem]{Definition}
\theoremstyle{remark}
\newtheorem{remark}[theorem]{Remark}

\newcommand{\R}{\mathbb{R}}
\newcommand{\N}{\mathcal{N}}
\newcommand{\ip}[2]{\langle #1, #2\rangle}
\newcommand{\relu}{\operatorname{relu}}
\newcommand{\reach}{\rightsquigarrow}

\title{Universality of certifiably 1-Lipschitz\\ ReLU networks}
\author{\authorname}
\date{}

\begin{document}
\maketitle

\begin{abstract}
Computing the Lipschitz constant of a neural network is NP-hard and intractable at practical
network sizes. Certifiably 1-Lipschitz ReLU networks are therefore typically trained under a
tractable layer-wise certificate: every weight matrix is constrained to have norm at most one.
Experiments and negative representation results suggested that this sufficient condition is
overly restrictive, motivating other activations and architectures under the same layer-wise
paradigm to improve performance. Yet none of these is known to be universal for vector-valued
maps with Euclidean norms. We prove that, in fact, plain ReLU networks under this constraint are
dense, uniformly on compact sets, in the 1-Lipschitz maps $\R^d\to\R^m$, for all $d$ and $m$. Our proof has been
formally verified in Lean~4 with Mathlib.
\end{abstract}

\section{Introduction}

Computing the Lipschitz constant of a neural network is NP-hard, even with two
layers~\cite{scaman2018lipschitz}, so constraining every layer to be 1-Lipschitz, i.e.\
non-expansive, is a standard way to obtain certifiably 1-Lipschitz models. With ReLU activations this has proved difficult in practice: many
normalization and orthogonalization schemes have been
proposed~\cite{cisse2017parseval,miyato2018spectral,trockman2021orthogonalizing,prach2022almost},
and alternative activations, such as GroupSort~\cite{anil2019sorting} and learnable linear
splines~\cite{neumayer2022approximation}, were introduced in part because ReLU appeared to limit
expressivity. Theory pointed the same way: with weights constrained in an operator $p$-norm,
$p>1$, ReLU networks cannot represent $x\mapsto|x|$
exactly~\cite{anil2019sorting,neumayer2022approximation}, and with $\infty$-norm constraints
there are 1-Lipschitz functions they cannot even approximate~\cite{neumayer2022approximation}.
Intuitively, ReLU trades slope for nonlinearity: to bend a map, some points must pass through
the flat part of the activation, where the slope drops to zero, and norm-constrained layers
cannot amplify afterwards to restore it.
Other than changing the activation, changing the architecture was proposed: 1-Lipschitz
residual ReLU networks were recently proved to be dense, uniformly on compact sets, in the
1-Lipschitz functions $\R^d\to\R$~\cite{murari2025approximation}, but this does not transfer to standard feed-forward networks or to vector-valued
outputs, a case that a recent survey describes as unknown~\cite{brau2026lipschitz}.
In sharp contrast with these experimental findings and partial theoretical results, we show that spectral-norm ($\ell^2$) constraints impose no limitation on expressivity in
the limit.
This does not contradict the experimental findings: our construction may require arbitrarily
large depth, and approximation is only one side of the story: it says nothing about how well such
networks can be trained.

Let $\N$ be the set of maps $A_L\circ\rho\circ A_{L-1}\circ\cdots\circ\rho\circ A_1$, where
$L\ge1$, each $A_j(x)=W_jx+b_j$ is affine between Euclidean spaces of arbitrary dimensions with
$\|W_j\|_2\le1$, and $\rho$ applies $\relu(t)=\max(t,0)$ to every coordinate.

\begin{theorem}\label{thm:main}
Let $K\subset\R^d$ be compact, let $f:\R^d\to\R^m$ be 1-Lipschitz on $K$, and let $\varepsilon>0$.
There is $N\in\N$ with $\|N(x)-f(x)\|_2<\varepsilon$ for all $x\in K$.
\end{theorem}

\begin{corollary}
A map $\R^d\to\R^m$ is a limit of networks in $\N$, uniformly on compact sets, if and only if
it is 1-Lipschitz.
\end{corollary}

\begin{proof}
Every $N\in\N$ is 1-Lipschitz, and 1-Lipschitz bounds pass to pointwise limits. The converse is
Theorem~\ref{thm:main}.
\end{proof}

Width and depth of the networks are not controlled; the construction uses both. The $\ell^2$
structure is essential: it rotates coordinates with orthogonal matrices.

Both statements are proved in Lean~4~\cite{lean4} with Mathlib~\cite{mathlib}, using only the
standard axioms; the code is available at \url{\repourl}. The proof below follows the
formalization.

\section{Proof}

\subsection{Overview}

We first give an informal version of the proof, to let the reader follow the argument; the
rigorous proof is deferred to the following subsections.

\begin{enumerate}
  \item \emph{Finite data.} It suffices to show that, for any finite set of samples $x_i\in K$,
    networks can map each $x_i$ arbitrarily close to $y_i=f(x_i)$: since both the network and
    $f$ are 1-Lipschitz, dense enough samples then give a uniform approximation on $K$
    (Section~\ref{sec:final}).
  \item \emph{Many small steps, each a proximal map.} We define an approximation map
    $\R^d\to\R^m$ as a chain of proximal maps of convex piecewise-linear functions $\varphi_k$,
    chosen as they can be approximated by networks (item~3), such that on the samples
    \[
      x_i\ \overset{L_{\mathrm{in}}}{\longmapsto}\ (x_i,0)
      \ \overset{\mathrm{prox}_{\varphi_1}}{\longmapsto}\ \cdots
      \ \overset{\mathrm{prox}_{\varphi_J}}{\longmapsto}\ (0,\,c^Jy_i)
      \ \overset{L_{\mathrm{out}}}{\longmapsto}\ c^Jy_i ,
    \]
    where $L_{\mathrm{in}}(x)=(x,0)$ and $L_{\mathrm{out}}(u,v)=v$ are linear maps with spectral
    norm one.
    \begin{itemize}
      \item \emph{Price of using proximal maps.} Unlike general 1-Lipschitz maps, proximal maps
        are firmly non-expansive: for any two samples, the difference $\Delta'$ after the move
        and the difference $\Delta$ before it satisfy $\|\Delta'\|^2\le\ip{\Delta'}{\Delta}$. The
        direct move $(x_i,0)\mapsto(0,y_i)$, although 1-Lipschitz on the samples, violates
        this, since $\Delta$ and $\Delta'$ are orthogonal.
      \item \emph{Small steps.} We therefore split it into $J$ steps: after $k$ steps, sample $i$
        sits at $\xi_i^k=(a_kx_i,\,b_ky_i)$, where $(a_k,b_k)=c^k(\cos k\theta,\sin k\theta)$
        turns by $\theta=\pi/(2J)$ and shrinks by $c=\cos\theta$ at each step. The shrink is a
        loss we would rather avoid, but it is necessary: when $\|y_i-y_j\|=\|x_i-x_j\|$, the step
        maps $\Delta$ to $\Delta'=s\,R_\theta\Delta$ with $R_\theta$ a rotation by $\theta$, so
        $\|\Delta'\|^2=s^2\|\Delta\|^2$ and $\ip{\Delta'}{\Delta}=s\cos\theta\,\|\Delta\|^2$,
        and firm non-expansiveness holds if and only if $s\le\cos\theta$.
      \item \emph{Each step is a proximal map.} Step $k$ is the proximal map of an explicitly
        defined function $\varphi_k$: the maximum of $|I|$ affine functions, one per sample
        $i\in I$, where the piece of sample $i$ has slope $\xi_i^{k-1}-\xi_i^k$, its targeted
        displacement during the step, and is active at $\xi_i^k$, its position after the step.
        Indeed, a proximal map is an implicit gradient step, $\mathrm{prox}_\varphi(v)=u$ if and
        only if $v-u\in\partial\varphi(u)$, so $\mathrm{prox}_{\varphi_k}(\xi_i^{k-1})=\xi_i^k$;
        that each piece is indeed active at its sample, i.e.\ no other piece lies above it there,
        is checked in Lemma~\ref{lem:finite}.
      \item \emph{Cost.} The shrink compounds to the total factor $c^J=\cos(\pi/2J)^J$, which
        tends to $1$ as $J\to\infty$ (Section~\ref{sec:finite}): smaller steps cost less.
    \end{itemize}
  \item \emph{From proximal maps to networks.} Each $\mathrm{prox}_{\varphi_k}$ is approximated by a
    ReLU network $N_k$ with weights of spectral norm at most one: the proximal map of a convex
    piecewise-linear function is close to the projection onto its epigraph, itself the limit of
    anchored iterations of halfspace projections, and each projection onto a halfspace
    $\{\ip uz\le\beta\}$, $\|u\|=1$, is exactly a network
    $z\mapsto Q^\top\sigma(Qz)$ with identity and clipping activations: an orthogonal $Q$ rotates $u$ onto the last coordinate axis,
    $\sigma$ applies $t\mapsto\min(t,\beta)$ to that coordinate and the identity
    to the others, and $Q^\top$ rotates back (Sections~\ref{sec:blocks} and~\ref{sec:step}).
    On any compact set, both activations are ReLUs up to sign flips and shifts, which the affine
    layers absorb without increasing the spectral norm; this is all the theorem needs: $\min(t,\beta)=\beta-\relu(\beta-t)$, and
    $t=\relu(t+B)-B$ for $t\ge-B$, which covers a compact set for $B$ large
    (Section~\ref{sec:final}). Since all blocks are 1-Lipschitz, the errors do not amplify, and the
    final network is $L_{\mathrm{out}}\circ N_J\circ\cdots\circ N_1\circ L_{\mathrm{in}}$.
\end{enumerate}

\subsection{Setting}

Throughout, all spaces are $\R^n$ with the Euclidean norm, and $\|W\|$ denotes the spectral
norm of a matrix $W$.

\begin{definition}[Network classes]
For a set $\mathcal{A}$ of functions $\R\to\R$, let $\N(\mathcal{A})$ be the smallest class of
maps $\R^p\to\R^q$ ($p,q\ge0$) that
\begin{itemize}
  \item contains the affine maps $x\mapsto Wx+b$ with $\|W\|\le1$,
  \item contains the coordinatewise maps $(z_i)_i\mapsto(\sigma_i(z_i))_i$ with all
    $\sigma_i\in\mathcal{A}$,
  \item is closed under composition of compatible maps: if $N_1:\R^p\to\R^q$ and
    $N_2:\R^q\to\R^r$ are in $\N(\mathcal{A})$, then so is $N_2\circ N_1:\R^p\to\R^r$.
\end{itemize}
\end{definition}

Since affine maps with $\|W\|\le1$ are closed under composition and $\rho\circ\rho=\rho$, we
have $\N(\{\relu\})=\N$. Most of the proof works with the identity and clipping activations,
\[
  \mathcal{C}=\{t\mapsto t\}\cup\{t\mapsto\min(t,c):c\in\R\},
  \qquad
  \N_{\mathcal C}=\N(\mathcal C),
\]
and Lemma~\ref{lem:relu} converts the result to ReLU. All activations in $\mathcal C$, as well
as $\relu$, are 1-Lipschitz, so every network in $\N_{\mathcal C}$ or in $\N$ is 1-Lipschitz.

\begin{definition}[Approximate reachability]
Let $I$ be a finite index set and let $P=(P_i)_{i\in I}$ and $Q=(Q_i)_{i\in I}$ be families of
points of $\R^p$ and $\R^q$, respectively. We write
\[
  P\reach Q
  \quad:\Longleftrightarrow\quad
  \forall\varepsilon>0\ \ \exists N\in\N_{\mathcal C}\ \ \forall i\in I:\ \
  \|N(P_i)-Q_i\|<\varepsilon ,
\]
and say that $Q$ is \emph{reachable} from $P$.
\end{definition}

Two elementary facts are used repeatedly:
\begin{itemize}
  \item if $N(P_i)=Q_i$ for all $i$ and some $N\in\N_{\mathcal C}$, then $P\reach Q$;
  \item $\reach$ is transitive: if $N_1(P_i)$ is $\varepsilon/2$-close to $Q_i$ and $N_2(Q_i)$
    is $\varepsilon/2$-close to $R_i$, then $N_2\circ N_1\in\N_{\mathcal C}$ maps $P_i$
    $\varepsilon$-close to $R_i$, because $N_2$ is 1-Lipschitz.
\end{itemize}

\subsection{Building blocks}\label{sec:blocks}

\begin{lemma}[Closure properties]\label{lem:closure}
Let $N,N_1,N_2\in\N_{\mathcal C}$.
\begin{enumerate}
  \item[(a)] For $s>0$, the map $x\mapsto sN(x/s)$ is in $\N_{\mathcal C}$.
  \item[(b)] If $N_1:\R^p\to\R^q$ and $N_2:\R^{p'}\to\R^{q'}$, then
    $(x,x')\mapsto(N_1x,N_2x')$ is in $\N_{\mathcal C}$.
  \item[(c)] If $N_1,N_2:\R^p\to\R^q$ and $\alpha\in(0,1)$, then
    $\alpha N_1+(1-\alpha)N_2\in\N_{\mathcal C}$.
\end{enumerate}
\end{lemma}

\begin{proof}
(a) By induction on the generation of $N$, using
\[
  s\bigl(W(x/s)+b\bigr)=Wx+sb,
  \qquad
  s\min(t/s,c)=\min(t,sc).
\]

(b) By induction, $(x,x')\mapsto(N_1x,x')$ is in $\N_{\mathcal C}$: an affine layer extended
by the identity still has spectral norm at most $1$, and an activation layer is extended by
identity activations. Doing the same in the second block and composing gives the claim.

(c) The map $x\mapsto(\sqrt\alpha\,x,\sqrt{1-\alpha}\,x)$ is a linear isometry. Follow it by
\[
  (u,v)\mapsto\Bigl(\sqrt\alpha\,N_1\bigl(u/\sqrt\alpha\bigr),\
  \sqrt{1-\alpha}\,N_2\bigl(v/\sqrt{1-\alpha}\bigr)\Bigr),
\]
which is in $\N_{\mathcal C}$ by (a) and (b), and then by
$(u,v)\mapsto\sqrt\alpha\,u+\sqrt{1-\alpha}\,v$, which has norm $1$ by Cauchy--Schwarz. The
composition is $\alpha N_1+(1-\alpha)N_2$.
\end{proof}

\begin{lemma}[Halfspace projections]\label{lem:proj}
Let $H_k=\{z:\ip{a_k}{z}\le b_k\}$, with $a_k\ne0$ and $k=1,\dots,M$, be halfspaces of $\R^n$
with nonempty intersection $C$. Let $P_k$ be the metric projection onto $H_k$ and
$T=P_M\circ\cdots\circ P_1$. Then:
\begin{enumerate}
  \item[(i)] $T\in\N_{\mathcal C}$;
  \item[(ii)] $\|Tz-p\|\le\|z-p\|$ for all $z\in\R^n$ and $p\in C$;
  \item[(iii)] if $Tz=z$, then $z\in C$.
\end{enumerate}
\end{lemma}

\begin{proof}
(i) It suffices to treat one $P_k$. Let $u=a_k/\|a_k\|$ and $c=b_k/\|a_k\|$. Then
\[
  P_k(z)=z-\ip uz\,u+\min(\ip uz,c)\,u ,
\]
which is the composition of three layers:
\begin{itemize}
  \item the isometry $z\mapsto(z-\ip uz\,u,\ \ip uz)\in\R^n\times\R$;
  \item the activation layer applying the identity to the first block and $\min(\cdot,c)$ to the
    last coordinate;
  \item the map $(v,t)\mapsto(v-\ip uv\,u)+tu$, whose norm is at most $1$.
\end{itemize}
Since the first block $z-\ip uz\,u$ is orthogonal to $u$, the last map returns
$z-\ip uz\,u+\min(\ip uz,c)\,u$.

(ii)--(iii) These are standard facts on compositions of projections (see, e.g.,~\cite{bauschke2017convex}); we include the short proof. Each $P_k$ is firmly non-expansive with respect to points of $H_k$:
\[
  \|P_kz-p\|^2+\|z-P_kz\|^2\le\|z-p\|^2
  \qquad(p\in H_k).
\]
Fix $p\in C$, put $z_0=z$ and $z_k=P_kz_{k-1}$. Summing these inequalities gives
\[
  \|Tz-p\|^2+\sum_{k=1}^M\|z_{k-1}-z_k\|^2\le\|z-p\|^2 .
\]
This proves (ii). If $Tz=z$, the sum vanishes, so all $z_k$ equal $z$, and $z=P_kz\in H_k$ for
every $k$.
\end{proof}

\begin{lemma}[Anchored fixed points]\label{lem:anchor}
Let $T:\R^n\to\R^n$ be 1-Lipschitz and let $p$ be a point with $\|Tz-p\|\le\|z-p\|$ for all
$z$. Let $w\in\R^n$ and $\alpha\in(0,1)$. The map $z\mapsto\alpha w+(1-\alpha)Tz$ is a
contraction, and its fixed point $z_\alpha$ satisfies
\begin{align}
  \ip{w-z_\alpha}{p-z_\alpha}&\le0, \label{eq:vi}\\
  \|z_\alpha-p\|&\le\|w-p\|, \label{eq:bd}\\
  (1-\alpha)\|z_\alpha-Tz_\alpha\|&\le2\alpha\|w-p\|. \label{eq:res}
\end{align}
Moreover, if $T\in\N_{\mathcal C}$, the maps
\[
  F_0=\mathrm{id},\qquad F_{r+1}(x)=\alpha x+(1-\alpha)\,T\bigl(F_r(x)\bigr)
\]
are networks in $\N_{\mathcal C}$, and $F_r(w)\to z_\alpha$ as $r\to\infty$.
\end{lemma}

This is the classical anchoring device of Halpern~\cite{halpern1967fixed}: for non-expansive
$T$, as $\alpha\to0$ the points $z_\alpha$ converge to the projection of $w$ onto the
fixed-point set of $T$ (Browder~\cite{browder1967convergence}). We only need the elementary estimates above and include
their short proof.

\begin{proof}
Write $z=z_\alpha$. From $z=\alpha w+(1-\alpha)Tz$ we get
\[
  (1-\alpha)(Tz-p)=(1-\alpha)(z-p)-\alpha(w-z).
\]
Squaring and using $\|Tz-p\|\le\|z-p\|$ gives
\[
  2\alpha(1-\alpha)\ip{z-p}{w-z}\ge\alpha^2\|w-z\|^2\ge0,
\]
which is \eqref{eq:vi}. It can be rewritten as $\|z-p\|^2\le\ip{z-p}{w-p}$, which gives
\eqref{eq:bd} by Cauchy--Schwarz. Finally $(1-\alpha)(z-Tz)=\alpha(w-z)$ and
$\|w-z\|\le\|w-p\|+\|z-p\|\le2\|w-p\|$, which gives \eqref{eq:res}.

The maps $F_r$ are in $\N_{\mathcal C}$ by Lemma~\ref{lem:closure}(c), and $F_r(w)$ is the
Picard iteration of the contraction started at $w$, so it converges to $z_\alpha$.
\end{proof}

\subsection{One proximal step}\label{sec:step}

\begin{lemma}[One proximal step]\label{lem:step}
Let $I$ be finite, and let $u_i,g_i\in\R^n$ and $\eta_i\in\R$ ($i\in I$) satisfy
\begin{equation}\label{eq:interp}
  \eta_j+\ip{g_j}{u_i-u_j}\le\eta_i\qquad\text{for all } i,j\in I.
\end{equation}
Then $(u_i+g_i)_{i\in I}\reach(u_i)_{i\in I}$.
\end{lemma}

\begin{remark}
Condition \eqref{eq:interp} says that $\varphi=\max_j\bigl(\eta_j+\ip{g_j}{\cdot-u_j}\bigr)$
is a convex function with $g_i\in\partial\varphi(u_i)$, so that
$\mathrm{prox}_\varphi(u_i+g_i)=u_i$. The proof approximates this proximal map on the data
through the projection onto the epigraph of $\delta\varphi$, for a small $\delta>0$.
\end{remark}

\begin{proof}
\emph{Step 1: the polyhedron.}
Adding the same constant to all $\eta_i$ preserves \eqref{eq:interp}, so we may assume
$\eta_i\ge0$. Let $\varepsilon>0$ and fix $\delta>0$ such that
\begin{equation}\label{eq:delta}
  \delta^2\bigl(\eta_i+\|g_i\|^2/4\bigr)\|g_i\|<\varepsilon/3\qquad\text{for all } i.
\end{equation}
In $\R^n\times\R$ consider the polyhedron
\[
  C=\bigl\{(z,t):\ t\ge0\ \text{ and }\ t\ge\delta\bigl(\eta_j+\ip{g_j}{z-u_j}\bigr)
  \text{ for all } j\bigr\},
\]
an intersection of halfspaces whose normals have last coordinate $-1$. By \eqref{eq:interp}
and $\eta_i\ge0$, the points $p_i=(u_i,\delta\eta_i)$ lie in $C$. Let $T$ be the composition of
the projections onto these halfspaces (Lemma~\ref{lem:proj}), and set
\[
  w_i=\bigl(u_i+g_i,\,-1/\delta\bigr).
\]

\emph{Step 2: limits of anchored fixed points.}
For $\alpha_k=1/(k+2)$, let $z_{k,i}$ be the fixed point of Lemma~\ref{lem:anchor} for $T$,
$p_i$, $w_i$ and $\alpha_k$. By \eqref{eq:bd}, $\|z_{k,i}\|\le\|w_i-p_i\|+\|p_i\|$, so along a
subsequence $z_{k,i}\to Z_i$ for all $i$. By \eqref{eq:res},
\[
  \|z_{k,i}-Tz_{k,i}\|\le\frac{2\|w_i-p_i\|}{k+1}\longrightarrow0,
\]
so $TZ_i=Z_i$, and $Z_i\in C$ by Lemma~\ref{lem:proj}(iii). Passing to the limit in
\eqref{eq:vi},
\begin{equation}\label{eq:vilim}
  \ip{w_i-Z_i}{p_i-Z_i}\le0 .
\end{equation}

\emph{Step 3: the epigraph estimate.}
Fix $i$ and write $Z_i=(z,t)$, $e=z-u_i$, $n=\|e\|$, $G=\|g_i\|$ and $\eta=\eta_i$. Since
$Z_i\in C$,
\[
  t\ge0
  \qquad\text{and}\qquad
  t-\delta\eta\ge\delta\ip{g_i}{e}.
\]
With $w_i-Z_i=(g_i-e,\,-1/\delta-t)$ and $p_i-Z_i=(-e,\,\delta\eta-t)$, inequality
\eqref{eq:vilim} reads
\begin{equation}\label{eq:expanded}
  n^2-\ip{g_i}{e}+(\delta^{-1}+t)(t-\delta\eta)\le0.
\end{equation}
Since $(\delta^{-1}+t)(t-\delta\eta)\ge(\delta^{-1}+t)\,\delta\ip{g_i}e
=\ip{g_i}e+t\delta\ip{g_i}e$, \eqref{eq:expanded} gives
\[
  n^2\le-t\delta\ip{g_i}e\le t\delta Gn,
  \qquad\text{hence}\qquad
  n\le t\delta G.
\]
Next, suppose $t>\delta(\eta+G^2/4)$. Then $t-\delta\eta>0$ and
\[
  (\delta^{-1}+t)(t-\delta\eta)\ge\delta^{-1}(t-\delta\eta)>G^2/4\ge Gn-n^2\ge\ip{g_i}e-n^2,
\]
contradicting \eqref{eq:expanded}. Hence $t\le\delta(\eta+G^2/4)$, and with \eqref{eq:delta},
\[
  \|z-u_i\|\le\delta^2\bigl(\eta_i+\|g_i\|^2/4\bigr)\|g_i\|<\varepsilon/3 .
\]

\emph{Step 4: the network.}
Choose $k$ in the subsequence with $\|z_{k,i}-Z_i\|<\varepsilon/3$ for all $i$, then $r$ with
$\|F_r(w_i)-z_{k,i}\|<\varepsilon/3$ for all $i$, where $F_r$ is built from $T$ and
$\alpha_k$ as in Lemma~\ref{lem:anchor}. Let $\pi_1(z,t)=z$ and
\[
  N(v)=\pi_1F_r(v,\,-1/\delta).
\]
Then $N\in\N_{\mathcal C}$, since its first layer is affine with isometric linear part. As
$N(u_i+g_i)=\pi_1F_r(w_i)$ and $\pi_1$ is 1-Lipschitz, $\|N(u_i+g_i)-u_i\|<\varepsilon$ for
all $i$.
\end{proof}

\subsection{Finite data}\label{sec:finite}

\begin{lemma}[Finite data]\label{lem:finite}
Let $I$ be finite and let $x_i\in\R^d$, $y_i\in\R^m$ satisfy $\|y_i-y_j\|\le\|x_i-x_j\|$ for
all $i,j\in I$. Then $(x_i)_{i\in I}\reach(y_i)_{i\in I}$.
\end{lemma}

\begin{proof}
For $P=(a,b)\in\R^2$ define the scaled graph
\[
  \Gamma(P)_i=(a\,x_i,\ b\,y_i)\in\R^{d+m},\qquad i\in I.
\]

\emph{Step 1: one step.}
Let $P=(a,b)$ and $Q=(a',b')$ satisfy
\[
  0\le a'\le a
  \qquad\text{and}\qquad
  \ip{Q}{Q-P}\le0,
\]
that is, $Q$ lies in the closed disc with diameter $[0,P]$. We claim
$\Gamma(P)\reach\Gamma(Q)$. Apply Lemma~\ref{lem:step} with
\[
  u_i=\Gamma(Q)_i,\qquad
  g_i=\Gamma(P)_i-\Gamma(Q)_i=\bigl((a-a')x_i,\,(b-b')y_i\bigr),\qquad
  \eta_i=\tfrac12\bigl(A\|x_i\|^2+B\|y_i\|^2\bigr),
\]
where $A=(a-a')a'\ge0$ and $B=(b-b')b'$. Note that $A+B=-\ip{Q}{Q-P}\ge0$. A direct
expansion gives
\begin{align*}
  \eta_i-\eta_j-\ip{g_j}{u_i-u_j}
  &=\tfrac12\bigl(A\|x_i-x_j\|^2+B\|y_i-y_j\|^2\bigr)\\
  &=\tfrac12\bigl(A(\|x_i-x_j\|^2-\|y_i-y_j\|^2)+(A+B)\|y_i-y_j\|^2\bigr)\ \ge\ 0,
\end{align*}
so \eqref{eq:interp} holds, and $u_i+g_i=\Gamma(P)_i$.

\emph{Step 2: the schedule.}
Fix $J\ge2$, let $\theta=\pi/(2J)$ and $c=\cos\theta$, and define
\[
  P_k=c^k\bigl(\cos k\theta,\ \sin k\theta\bigr),\qquad 0\le k\le J .
\]
Then $P_{k+1}$ is $P_k$ rotated by $\theta$ and scaled by $c$, so
\[
  \ip{P_{k+1}}{P_k}=c^{2k+1}\cos\theta=\|P_{k+1}\|^2,
  \qquad\text{i.e.}\qquad
  \ip{P_{k+1}}{P_{k+1}-P_k}=0 .
\]
Moreover $0\le c^{k+1}\cos((k+1)\theta)\le c^k\cos(k\theta)$, since $(k+1)\theta\le\pi/2$.
By Step~1 and transitivity, $\Gamma(P_0)\reach\Gamma(P_J)$, where
\[
  P_0=(1,0),\qquad P_J=(0,c^J).
\]

\emph{Step 3: conclusion.}
The linear map $x\mapsto(x,0)$ sends $x_i$ to $\Gamma(P_0)_i$, and the linear map
$(u,v)\mapsto v$ sends $\Gamma(P_J)_i$ to $c^Jy_i$; both have norm at most $1$. Hence
$(x_i)\reach(c^Jy_i)$. Since $\theta\le\pi/4$, Bernoulli's inequality gives
\[
  1\ge c^J\ge\bigl(1-\theta^2/2\bigr)^J\ge1-\frac{\pi^2}{8J},
  \qquad\text{so}\qquad
  \|c^Jy_i-y_i\|\le\frac{\pi^2}{8J}\max_j\|y_j\|,
\]
which is arbitrarily small for $J$ large.
\end{proof}

\subsection{From clipping to ReLU, and proof of the theorem}\label{sec:final}

\begin{lemma}[ReLU suffices on compact sets]\label{lem:relu}
For every $N\in\N_{\mathcal C}$ and every compact set $K$, there is $N'\in\N$ with $N'=N$ on
$K$.
\end{lemma}

\begin{proof}
By induction on the generation of $N$. Affine layers are already in $\N$.

\emph{Activation layers.} Let $\sigma=(\sigma_i)$ with $\sigma_i\in\mathcal C$. Take $B\ge0$
with $\|z\|\le B$ on $K$, so that $z_i\ge-B$ on $K$. Let $D$ be the diagonal matrix with
$D_{ii}=1$ if $\sigma_i$ is the identity and $D_{ii}=-1$ if $\sigma_i=\min(\cdot,c_i)$, and
let
\[
  (\beta_i,\beta'_i)=
  \begin{cases}
    (B,\,-B) & \text{if $\sigma_i$ is the identity},\\
    (c_i,\,c_i) & \text{if $\sigma_i=\min(\cdot,c_i)$}.
  \end{cases}
\]
Then $z\mapsto D\rho(Dz+\beta)+\beta'$ is in $\N$ and equals $\sigma$ on $K$, because
\[
  \relu(t+B)-B=t\quad\text{for } t\ge-B,
  \qquad
  c-\relu(c-t)=\min(t,c)\quad\text{for all } t.
\]

\emph{Compositions.} For $N=N_2\circ N_1$, take $N_1'$ agreeing with $N_1$ on $K$ and $N_2'$
agreeing with $N_2$ on the compact set $N_1(K)$.
\end{proof}

\begin{proof}[Proof of Theorem~\ref{thm:main}]
We may assume $K\ne\emptyset$. Let $\{x_i\}_{i\in I}\subset K$ be a finite $\varepsilon/3$-net
of $K$ and $y_i=f(x_i)$. Since $f$ is 1-Lipschitz on $K$, Lemma~\ref{lem:finite} gives
$N_0\in\N_{\mathcal C}$ with $\|N_0(x_i)-y_i\|<\varepsilon/3$ for all $i$. For $x\in K$ pick
$i$ with $\|x-x_i\|<\varepsilon/3$. Since $N_0$ and $f|_K$ are 1-Lipschitz,
\[
  \|N_0(x)-f(x)\|
  \le\|N_0(x)-N_0(x_i)\|+\|N_0(x_i)-y_i\|+\|f(x_i)-f(x)\|
  <\varepsilon .
\]
Finally, by Lemma~\ref{lem:relu}, some $N\in\N$ equals $N_0$ on $K$.
\end{proof}

\section*{Use of AI tools}

The construction, the Lean~4 formalization and a first draft of this manuscript were produced
with assistance from a large language model.

\section*{Code availability}

The Lean~4 source (Lean and Mathlib v4.33.0) is available at \url{\repourl}. The trusted
definitions are the network class (\texttt{IsNetOf}, \texttt{IsReLU} in \texttt{Network.lean})
and the two theorems in \texttt{Main.lean}; \texttt{Check.lean} prints their axioms.

\end{document}